\documentclass[runningheads]{llncs}

\usepackage{eccv}

\usepackage{eccvabbrv}

\usepackage{graphicx}
\usepackage{booktabs}

\usepackage[accsupp]{axessibility}  

\usepackage{multirow}
\usepackage{wrapfig}

\usepackage{xurl} 
\usepackage{hyperref}

\usepackage{orcidlink}

\usepackage[shortlabels]{enumitem}

\begin{document}

\title{Pixel-wise Geo-registration of Drone  Images} 
\titlerunning{Pixel-wise Geo-registration of Drone  Images}


\author{Qingyang Liu\textsuperscript{*} \quad David G. Shatwell\textsuperscript{*} \quad Parth Parag Kulkarni \quad Mubarak Shah}

\authorrunning{Q.~Liu et al.}

\institute{
Institute of Artificial Intelligence, University of Central Florida, USA\\
}

\maketitle

\begingroup
\renewcommand{\thefootnote}{\fnsymbol{footnote}}
\footnotetext[1]{These authors contributed equally to this work.}
\endgroup

\begin{abstract}

Cross-view geo-registration is the task of aligning a query image to a geodetically accurate reference (e.g., satellite image), so that each query pixel maps to a real-world GPS coordinate. 
Most existing work addresses the related problem of cross-view geo-localization, where the goal is typically to estimate the camera center using retrieval, classification, matching, or regression. Because these approaches do not provide dense supervision, they are poorly suited for learning and evaluating pixel-wise alignment.
We introduce \emph{SkyReg}, a geometry-aware geo-registration model that estimates the transformation 
between the query and reference images by explicitly modeling the 3D scene geometry. Applying this transformation to warp the query into the reference frame yields pixel-wise geo-localization, without relying on 2D point matches, and remains robust to occlusions and large viewpoint changes. To enable training and standardized evaluation, we release (i) \emph{SkyReg-Train}, a large-scale dataset of drone--satellite images annotated with per-pixel GPS coordinates, depth maps, and camera parameters derived from LiDAR and structure-from-motion, and (ii) \emph{SkyReg-Bench}, a held-out benchmark of unseen Urban and Suburban scenes with the same dense annotations. SkyReg achieves state-of-the-art performance against strong retrieval and homography baselines, demonstrating the value of geometry-aware models and dense geodetic benchmarks for cross-view geo-registration. 
Dataset available at
\href{https://parthpk.github.io/skyreg-webpage}{%
  \textcolor{cyan!60!blue}{\nolinkurl{https://parthpk.github.io/skyreg-webpage}}%
}.

\end{abstract}

\section{Introduction}

Cross-view geolocalization is the task of estimating the GPS coordinates of a query image (typically captured from a ground or drone viewpoint) by associating it with a geo-referenced satellite view. Existing methods generally follow two paradigms: \emph{retrieval}, where given a query image the goal is to identify the best-matching satellite image from a large gallery of geo-tagged images \cite{zhu2022transgeo, Xia_2025_CVPR, zhu2021vigor}, and \emph{3-DOF pose regression}, where the reference is a single satellite image and the objective is to predict the query's location and compass orientation in the satellite image plane \cite{Xia_2025_CVPR, 10373898}. Both paradigms, however, ultimately reduce the problem to estimating the camera geo-location.

This image-level formulation does not capture the spatial structure within a scene. In many practical settings, knowing the camera position alone is insufficient: a small localization error can translate into much larger spatial displacement for distant structures, especially in urban environments with significant depth variation and oblique viewpoints. Applications such as infrastructure inspection, urban planning, and disaster assessment require reasoning about the geographic location of \emph{visible scene content}, not just the camera center. Related tasks, including map updating and defense-oriented geospatial analysis often demand precise geo-referencing of specific targets within the image. These requirements motivate a shift from image-level localization to \emph{pixel-level geo-registration}, where each query pixel is mapped to real-world latitude and longitude.

Geo-registration has been studied extensively in remote sensing, but most existing approaches are tailored to small viewpoint changes, assume approximately planar geometry, and rely on homography-based alignment \cite{ozcanli2014automatic, berton2024earthmatch, yuan2020automated}. Evaluation is also often limited to sparse correspondences rather than dense, per-pixel geodetic consistency. As a result, these methods do not adequately address satellite--drone geo-registration, where large viewpoint changes and heterogeneous sensing modalities (orthographic satellite imagery versus perspective drone imagery) make the problem substantially more challenging. Moreover, the field lacks standardized training data and benchmark protocols for assessing fine-grained cross-view geo-registration across methods.

In this paper, we propose \emph{SkyReg}, a drone geo-registration method that leverages state-of-the-art 3D reconstruction backbones to model scene geometry and predict pixel-wise GPS coordinates, even under occlusions and large viewpoint differences. Unlike previous approaches that rely on homography estimation or assume known satellite DEMs, SkyReg uses camera parameters and dense point maps for the query and reference images. It then computes a query-to-reference transformation by composing the geometric mappings that lift each image plane into a shared 3D reference frame. The resulting warp projects query pixels into the reference image plane, where geodetic coordinates are assigned by interpolating the reference latitude–longitude map.

We also introduce \emph{SkyReg-Train}, a diverse training set of drone images and geo-referenced satellite views, covering multiple scene types and acquisition conditions. Each image is annotated with dense per-pixel latitude--longitude labels, along with depth maps and camera parameters used as training signals, which are derived from LiDAR and structure-from-motion (SfM). For quantitative evaluation, we propose \emph{SkyReg-Bench}, a held-out benchmark of unseen Urban and Suburban scenes with the same pixel-level GPS annotations, designed to evaluate models under challenging distribution shifts. We evaluate a broad suite of baselines, including feature matching and homography-based alignment. Across all settings, we observe a substantial gap between image-level localization and dense geo-registration, underscoring the need for geometry-aware cross-view methods tailored to pixel-wise alignment.

\noindent To summarize, our contributions are the following:
\begin{itemize}
    \item We introduce \emph{SkyReg}, the first 3D-aware drone--satellite geo-registration method that explicitly models the 3D scene geometry to infer pixel-wise GPS coordinates, remaining robust under occlusions and large viewpoint changes.
    \item We release \emph{SkyReg-Train}, a large-scale training dataset for drone geo-registration, providing dense per-pixel latitude--longitude labels together with depth maps and camera parameters across diverse scenes.
    \item We propose \emph{SkyReg-Bench}, a held-out benchmark for evaluating generalization to unseen Urban and Suburban environments, multiple reference modalities (orthorectified and perspective), and varied scene layouts.
    \item We formalize pixel-wise geo-registration with standardized evaluation protocols, metrics, and benchmark a broad set of retrieval- and homography-based baselines, showing consistent improvements in geodetic error and recall.
\end{itemize}

\vspace{-1em}

\section{Background and Related Work}

\subsection{Geolocalization}

Geo-localization is the task of inferring \emph{where} an image was captured by estimating its GPS coordinates. Existing approaches can be broadly grouped into four directions. (i) \textbf{Global Level Classification} methods \cite{weyand2016planet, seo2018cplanet, vo2017revisiting, muller2018geolocation, pramanick2022world, kulkarni2024cityguessr, clark2023we} divide the Earth into discrete geo-cells and train a model to predict the corresponding location class. These methods require only a single forward pass at inference time, but their accuracy is inherently limited by the geo-cell resolution. (ii) \textbf{City Level Classification} methods \cite{Trivigno_2023, Xu_2024, Xu_2025}, similar to Global Level Classification, try to classify input into discrete geo-cells, except the geo-cell gallery are only at a city level instead of global level, thus allowing a more fine grained classification. (iii) \textbf{Retrieval} methods predict location by matching a query image to a gallery of geo-tagged references and transferring the retrieved reference coordinates to the query. Galleries may consist of GPS embeddings \cite{vivanco2023geoclip, klemmer2025satclip, shatwell2025gt, shatwell2026tiger, kulkarni2026vidtag}, ground view images (same-view)\cite{berton2025megaloc, izquierdo2024optimal, tzachor2024effovpr, arandjelovic2016netvlad, ali2023mixvpr, berton2023eigenplaces, keetha2023anyloc}, or top-down satellite images (cross-view) \cite{regmi2019bridging, shi2019spatial, zhu2022transgeo, toker2021coming, shi2020looking, zhu2021vigor, zheng2020university, li2024unleashing, shugaev2024arcgeo, ye2024cross}. Cross-view methods like TransGeo~\cite{zhu2022transgeo} learn a shared embedding space for ground-level and satellite imagery with re-ranking for refining nearest-neighbor retrieval, while University-1652 \cite{zheng2020university} jointly embeds ground, drone, and satellite views into a unified feature space. (iv) \textbf{Fine-grained cross-view localization} methods \cite{zhu2021vigor, ye2025cross, wang2024view, Xia_2025_CVPR, 10.1007/978-3-031-19842-7_6, 10373898} go a step further by estimating the camera location within a geodetically accurate reference image (e.g., satellite), often leveraging an approximate GPS prior to restrict the search region.

Although these approaches can accurately recover the camera’s geo-location and sometimes the 3-DoF camera orientation, they operate at the image level in the 2D domain and therefore do not provide pixel-wise geographic correspondence. Estimating the camera center reveals the observer’s position, but it does not determine where a visible object (e.g., a building facade or vehicle) lies in real-world GPS coordinates. Unlike geo-localization, which assigns a single coordinate pair to the entire image, geo-registration estimates coordinates for each pixel, enabling dense geographic alignment.

\vspace{-1em}

\subsection{Georegistration}

Geo-registration aligns a query image to an Earth-fixed coordinate system (e.g., WGS84/UTM) so that image pixels can be mapped to geodetic coordinates. Given a query image \(I_q(u,v)\) and a geodetically accurate reference \(I_{\rm ref}(x,y)\), the goal is to estimate a mapping \(T:(u,v)\mapsto(x,y)\) such that \(I_q(u,v)\approx I_{\rm ref}(T(u,v))\).

Early work primarily treated geo-registration as pose refinement: starting from a noisy initialization, methods iteratively improved alignment using DEM-based rendering and template/feature matching \cite{sheikh, sheikh2004feature}, or stabilized aerial video by remapping frames with planar homographies \cite{hafiane2008uav}, which can be brittle in non-planar scenes with strong 3D structure. Later approaches include joint correction across multiple satellite images via probabilistic 3D surface reconstruction \cite{ozcanli2014automatic} and learned cross-view descriptors followed by feature matching \cite{yuan2020automated}, but many formulations still rely on 2D homography estimation and therefore struggle under large parallax and viewpoint changes.

Recent progress in learned correspondence has improved robustness in challenging regimes: SuperPoint/SuperGlue \cite{detone2018superpoint, sarlin2020superglue} and RoMa \cite{edstedt2024roma} provide stronger sparse and dense matches, and have been adopted in modern homography-based registration pipelines \cite{berton2024earthmatch}. However, many systems still default to planar warps (e.g., homographies) when aligning cross-view imagery, and recent 3D rendering-based pipelines \cite{bredvik} typically require multiple drone views to build consistent geometry. In contrast, we target geo-registration from a \emph{single} drone image to a geodetically accurate satellite reference, motivating geometry-aware, feed-forward inference beyond 2D alignment.

\vspace{-1em}

\subsection{3D Reconstruction and Multi-View Geometry}

Models utilizing 3D have become increasingly prominent~\cite{meng2024}.  
Recent feed-forward reconstruction models such as DUSt3R \cite{dust3r}, MASt3R \cite{mast3r}, and VGGT \cite{vggt} advance multi-view geometry by predicting dense \emph{point maps} and \emph{camera parameters} from a sparse set of images. They recover 3D structure in a shared coordinate frame (up to a global similarity transform) without running a full SfM pipeline. However, these models often degrade under the extreme appearance and viewpoint gaps of satellite--drone alignment: satellite imagery differs in scale, projection (near-orthographic), and radiometry, and the baseline to oblique drone views is typically outside the regime covered by standard reconstruction benchmarks. Consequently, predicted point maps and camera poses can become unreliable, limiting direct use for satellite-conditioned geo-registration.

Aerial MegaDepth \cite{vuong2025aerialmegadepth} partially bridges this gap by providing aerial scenes with calibrated intrinsics, poses, and depth in a unified frame. While valuable for cross-view reconstruction, it does not target the satellite setting or explicitly benchmark pixel-level geo-registration.

Overall, existing benchmarks and methods largely operate at the image-level (camera-center localization) or rely on planar homographies and/or multi-view inputs, which are brittle under the parallax, occlusions, and modality gap of satellite--drone pairs. Meanwhile, feed-forward 3D reconstruction is promising but lacks satellite-conditioned training data and standardized pixel-level evaluation. This motivates our geometry-aware, single-image geo-registration model and the accompanying dataset and benchmark with dense geodetic supervision.

\section{Method}

SkyReg operates in two stages. First, a feed-forward 3D reconstruction backbone predicts dense point maps, camera intrinsics, and the relative pose between the query and reference views. Second, we compose these predictions into a query-to-reference warping function that lifts query pixels into 3D and reprojects them onto the reference image plane. We then obtain a GPS coordinate for each query pixel by interpolating the reference’s geodetic map at the warped locations.

\vspace{-1em}

\subsection{Inputs}

The inputs to our geo-registration framework consist of a georeferenced reference image $I_{r}(\mathbf{x}_{r})$ and an uncalibrated query image $I_{q}(\mathbf{x}_{q})$, where $\mathbf{x} = [u,v]^\top$ corresponds to pixel coordinates. The query image $I_{q}(\mathbf{x}_{q})$ is captured from a low altitude, often at an oblique angle, and lacks any associated metadata. Its only available information is the pixel intensity values. The reference image, typically a satellite view, has known intrinsics and pose $(R^{r\rightarrow w}, \mathbf{t}^{r\rightarrow w})$ relative to an Earth-Centered, Earth-Fixed (ECEF) world coordinate system. This image may optionally include a Digital Elevation Map (DEM) $D(\mathbf{x}_{r})$ that provides height information for each pixel. In addition, we assume that each pixel in the reference image and DEM can be mapped to a 3D location on the surface of the WGS-84 ellipsoid in the ECEF coordinate system, such that $\mathbf{X}_{\mathrm{WGS84}} = \mathcal{E}(\mathbf{x}_{r}) = [X_{\mathrm{WGS84}},Y_{\mathrm{WGS84}},Z_{\mathrm{WGS84}}]^\top$. Each 3D point $\mathbf{X}_{\mathrm{WGS84}}$ corresponds to latitude-longitude pairs $(\varphi, \lambda)$ described by:

\begin{equation}
\begin{aligned}
e^2 &= 1-\frac{b^2}{a^2}, \quad
N(\varphi) = \frac{a}{\sqrt{1-e^2\sin^2\varphi}}, \\
X &= N(\varphi)\cos\varphi\cos\lambda, \\
Y &= N(\varphi)\cos\varphi\sin\lambda, \\
Z &= (1-e^2)N(\varphi)\sin\varphi,
\end{aligned}
\label{eq:geodetic2ecef}
\end{equation}

\noindent where $a$ and $b$ are the WGS84 ellipsoid semi-major and semi-minor axes.

We aim to \textbf{georegister} the query image by finding the warping function $\mathbf{x}_{q}' = \mathrm{warp}(\mathbf{x}_{q})$ that maps pixel coordinates between the two image planes. Unlike traditional planar homography-based approaches, our formulation explicitly models 3D geometry and does not assume a flat ground plane, making it suitable for low-altitude images with noticeable parallax and terrain relief. Once the warping function is estimated, we can assign a geographic coordinate (latitude and longitude, or equivalently $\mathbf{X}_{\mathrm{WGS84}}$) to every query pixel based on their warped location in the reference image. Additionally, we can project the query image onto the geodetically reference image and create a mosaic.

\vspace{-1em}

\subsection{Feed-Forward 3D Reconstruction}

Our method begins by estimating the geometric relationship between the query and reference images using a feed-forward multi-view 3D reconstruction network (i.e., MASt3R \cite{mast3r}). Given $I_{q}$ and $I_{r}$, the model predicts two sets of point-maps in two forward passes: $(X^{r,r}, X^{q,r})$ and $(X^{q,q}, X^{r,q})$, where $X^{i,j}$ corresponds to the point-map of image $i$ in camera's $j$ reference frame, and each pixel in point-map $X^{i,j}$ is a 3D vector.

After predicting the point-maps, we can estimate the intrinsics of the query image $K_{q}$ and reference image $K_{r}$, as well as the relative rotation $R^{q \rightarrow r}$, translation $\mathbf t^{q \rightarrow r}$, and scale $\sigma^{q \rightarrow r}$. Following DUSt3R~\cite{dust3r}, the intrinsics are recovered via solving an optimization problem. We assume that the principal point $\mathbf{c}$ is centered and solve for the query focal length $f^*_{q}$ as follows:

\begin{equation}
    \arg \min_{f_{q}} \sum_{\mathbf{x}_q} \left|\left| (\mathbf{x}_q - \mathbf{c}) - f_{q} \frac{X^{q,q}_{0:1}(\mathbf{x}_q)}{X^{q,q}_{2}(\mathbf{x}_q)} \right|\right|,
\end{equation}

\noindent where $X^{q,q}_{k}(\mathbf{x}_q)$ represents the $k$th channel of the point map, \textit{i.e.}, the $X$, $Y$, or $Z$ coordinates of the 3D point at pixel location $\mathbf{x}_q$. This process can also be performed similarly for the reference image to obtain $f^*_{r}$ and $K_{r}$. Similarly, we can recover the relative transformation $\hat P^{q \rightarrow r}$ through Procrustes alignment:

\begin{equation}
    \arg \min_{\hat P^{q \rightarrow r}} \sum_{\mathbf{x}_q} \left|\left| h^{-1}(\hat P^{q \rightarrow r} h(X^{q,q}(\mathbf{x}_q))- X^{q,r}(\mathbf{x}_q)\right|\right| ^2,
\end{equation}

\begin{equation}
    \hat P^{q \rightarrow r} = \begin{bmatrix}
        \sigma^{q \rightarrow r} R^{q \rightarrow r} & \mathbf{t}^{q \rightarrow r} \\
        \mathbf{0}^\top & 1
    \end{bmatrix},
\end{equation}

\noindent where $h(\cdot)$ transforms the 3D point into homogeneous coordinates.

\vspace{-1em}

\subsection{Pixel-Wise Geo-Registration}

In order to geo-register the query image, we need to find the warping function that maps pixel coordinates between the query and reference image planes, such that $\mathbf{x}_q' = \mathrm{warp}(\mathbf{x}_q)$. We can easily create this function from the predicted query point-map. Since $X^{q,r}$ is already in the reference coordinate frame, we simply project the 3D points back into the reference image plane using $K_r$. Mathematically, we define it as:
\begin{equation}
\label{eq:warp}
\mathbf{x}_q' = \mathrm{warp}(\mathbf{x}_q)
= h^{-1}\!\left(K_r X^{q,r}(\mathbf{x}_q)\right),
\end{equation}

Using this warping function, we can easily find the global 3D location (or alternatively, the latitude and longitude) of each query pixel simply by finding its location in the reference image plane passing it to the ellipsoid mapping function $\mathcal{E}(\cdot)$:

\begin{equation}
    \mathbf{X}_{\mathrm{WGS84}} = \mathcal{E}(\mathbf{x}_q') = \mathcal{E}(\mathrm{warp}(\mathbf{x}_q)) = \mathcal{E}(h^{-1}(K_r X^{q,r}(\mathbf{x}_q))).
\end{equation}

Alternatively, we can create a mosaic of the query image overlaid on top of the reference by assigning the RGB values of the query to the warped pixel coordinates in the reference image space.

\vspace{-1em}

\subsection{Training Details}

We instantiate our feed-forward geo-registration model from the MASt3R \cite{mast3r} architecture and train it on the SkyReg-Train dataset, initializing from the checkpoint pretrained on AerialMegaDepth \cite{vuong2025aerialmegadepth}. During training, we feed the query image and the geo-referenced reference image as a pair into the 3D backbone. For each forward pass, the network predicts two dense point maps, both expressed in the reference camera coordinate frame, together with per-pixel confidence maps. As supervision, we construct ground-truth point maps from the depth maps and camera parameters. Specifically, for a pixel $(i,j)$ in view $n$, its 3D point expressed in the coordinate frame of view $m$ is computed as
\begin{equation}
X_{i,j}^{n,m} = P_m P_n^{-1}\, h\!\left(K^{-1}[iD_{i,j},\, jD_{i,j},\, D_{i,j}]^\top\right),
\end{equation}
where $K$ is the intrinsic matrix, $D_{i,j}$ is the depth at pixel $(i,j)$, $P_n$ and $P_m$ denote the camera extrinsic matrices, and $h(\cdot)$ is the homogeneous lifting operator.

We supervise the predicted point maps using the confidence-weighted regression loss introduced in DUSt3R~\cite{dust3r}. The loss is defined as
\begin{equation}
\mathcal{L}_{\mathrm{conf}}
=
\sum_{v \in \{1,2\}}
\sum_{i \in \mathcal{D}^{v}}
C_{i}^{v,1}\,\ell_{\mathrm{regr}}(v,i)
\;-\;
\alpha \log C_{i}^{v,1},
\end{equation}
where
\begin{equation}
\ell_{\mathrm{regr}}(v,i)
=
\left\lVert
\frac{1}{z} X_{i}^{v,1}
-
\frac{1}{z} \bar{X}_{i}^{\bar{v},1}
\right\rVert.
\end{equation}
Here, $X_{i}^{v,1}$ denotes the predicted 3D point, $\bar{X}_{i}^{\bar{v},1}$ is the corresponding ground-truth point, $C_{i}^{v,1}$ is the predicted confidence, $\mathcal{D}^{v}$ is the set of valid supervised pixels in view $v$, and $\alpha$ controls the confidence regularization term.

We train the model for 20 epochs. To preserve the strong geometric priors of the 3D backbone while adapting it to the cross-view geo-registration setting, each epoch is formed by a balanced mixture of training pairs from multiple sources. In particular, every epoch contains 20k randomly sampled pairs from Urban scenes, 20k pairs from Landmarks scenes, and 20k pairs from traditional 3D reconstruction datasets, including ARKitScenes \cite{baruch2021arkitscenes}, ScanNet++ \cite{yeshwanth2023scannet++}, and CO3D \cite{reizenstein2021common}.

We optimize the network with AdamW\cite{loshchilov2017decoupled} using a base learning rate of 1e-4, weight decay of 0.05, and momentum parameters $(\beta_1,\beta_2)=(0.9, 0.95)$. We use an effective batch size of 48 image pairs, train with input resolutions between 512$\times$160 and 512$\times$384. The learning rate is scheduled using a cosine decay with one warm-up iteration. Training is performed on three GPUs for approximately 20 hours. The models are trained on NVIDIA H100 GPUs, and inference is conducted on a single NVIDIA RTX 3090 GPU.

\section{Dataset}

We introduce \emph{SkyReg}\footnote{https://github.com/dshatwell23/skyreg-dataset}, a dataset for drone--satellite geo-registration, with \emph{SkyReg-Train} and \emph{SkyReg-Bench} as its training and benchmarking components respectively. SkyReg spans diverse locations, scene types, and camera configurations, and provides dense supervision, including per-pixel GPS coordinates, metric depth, and full 6-DoF camera poses, which enable evaluation beyond image-level geo-localization.

Each sample contains a geodetically accurate reference image $I_r(\mathbf{x}_r)$ with a per-pixel latitude--longitude array. We support two reference modalities: (i) orthorectified satellite tiles paired with a digital elevation map $DEM(\mathbf{x}_r)$, and (ii) perspective-projection satellite views paired with metric depth $D_r(\mathbf{x}_r)$ and camera parameters $(K_r, R^{w\rightarrow r}_r, \mathbf{t}^{w\rightarrow r}_r)$ defined in an ECEF world frame. The query is always a perspective-projection drone image $I_q(\mathbf{x}_q)$ with metric depth $D_q(\mathbf{x}_q)$ and camera parameters $(K_q, R^{w\rightarrow q}_q, \mathbf{t}^{w\rightarrow q}_q)$, also in ECEF.

\begin{figure}[t]
    \centering
    \includegraphics[width=0.85\linewidth]{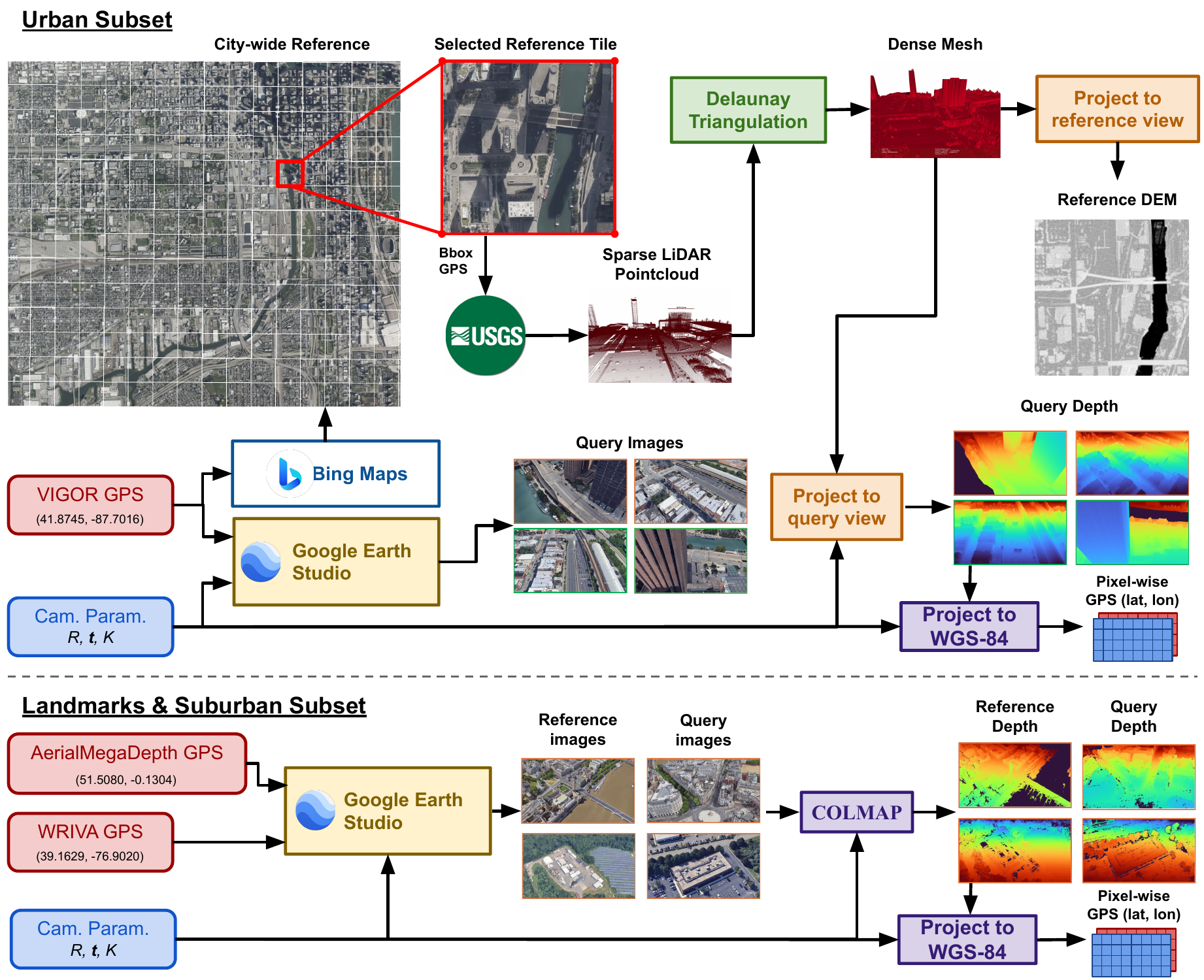}
    \caption{\textbf{SkyReg data construction for SkyReg-Train and SkyReg-Bench.}
    \textbf{Top (Urban).} We use VIGOR~\cite{zhu2021vigor} GPS coordinates from three cities (San Francisco, Seattle, Chicago) to download orthorectified Bing Maps tiles and render perspective drone views in Google Earth Studio. LiDAR point clouds from USGS are densified via Delaunay triangulation and projected into the known cameras to produce DEMs, depth maps, and per-pixel latitude--longitude labels through projection to the WGS-84 ellipsoid.
    \textbf{Bottom (Landmarks \& Suburban).} Landmark drone images come from AerialMegaDepth~\cite{vuong2025aerialmegadepth}, while landmark references and all Suburban imagery are rendered in Google Earth Studio. Depth is obtained with a structure-from-motion pipeline using fixed camera parameters, and pixel-wise GPS is computed using the same 3D-to-WGS-84 projection as in Urban.}
    \label{fig:dataset}
\end{figure}

\begin{figure}[t]
    \centering
    \includegraphics[width=0.95\linewidth]{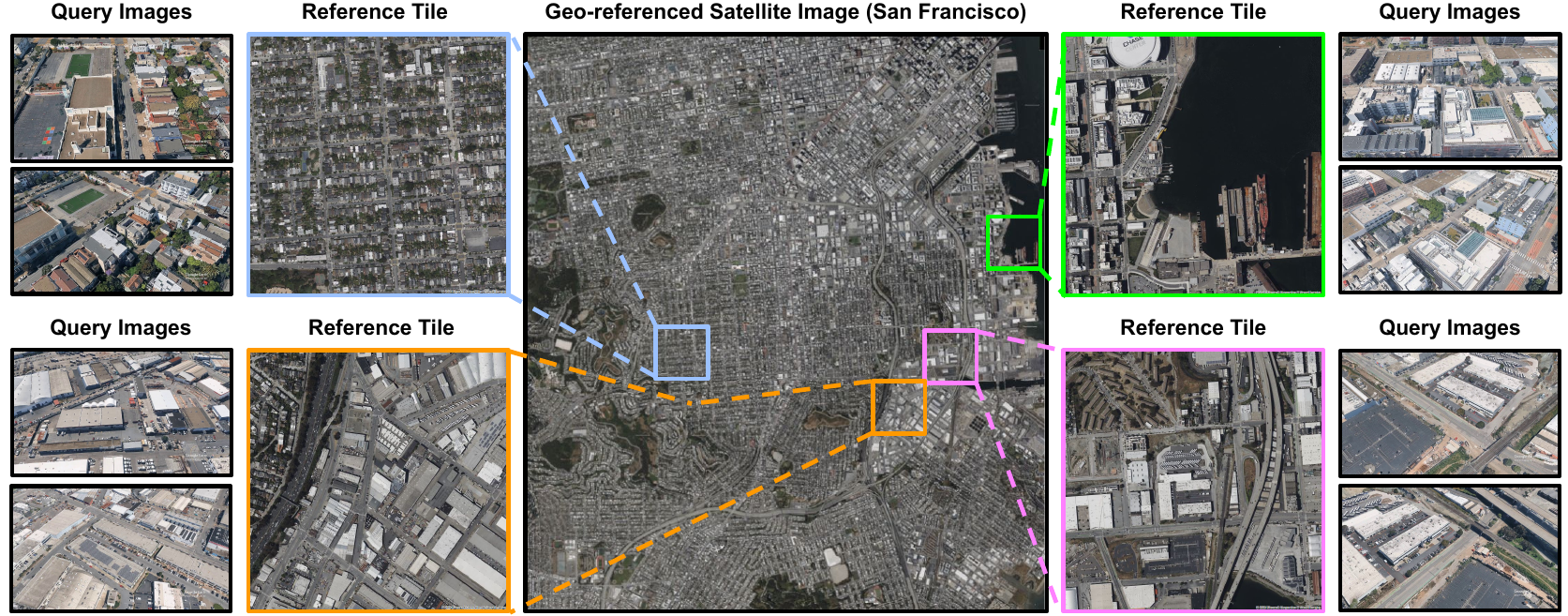}
    \caption{\textbf{Representative samples from SkyReg-Train (Urban).} The center panel shows a large orthorectified satellite image of San Francisco. We create reference satellite images from local reference tiles. The left/right panels show example drone queries associated with each reference tile.}
    \label{fig:dataset_examples}
\end{figure}

We aggregate data from multiple sources into three subsets: \textbf{Urban}, \textbf{Landmarks}, and \textbf{Suburban}. (i) \textbf{Urban} is built from three U.S. cities (Chicago, San Francisco, and Seattle) by sampling GPS locations to match the coordinate distribution of VIGOR \cite{zhu2021vigor}. (ii) \textbf{Landmarks} comprises 83 scenes centered on well-known landmarks worldwide, leveraging drone imagery from AerialMegaDepth \cite{vuong2025aerialmegadepth}. (iii) \textbf{Suburban} follows the GPS distribution of drone images in the WRIVA Public dataset \cite{wrivapublic}. Urban satellite references are orthorectified Bing Maps tiles, while the drone images are rendered from Google Earth. The Landmarks and Suburban subsets use perspective-projection drone and satellite views also rendered in Google Earth. We derive metric depth maps for Urban scenes from USGS LiDAR, and for Landmarks/Suburban we run COLMAP with fixed camera parameters, optimizing only for metric depth. Figures \ref{fig:dataset} and \ref{fig:dataset_examples} show the data collection pipeline and sample images from the Urban subset, respectively. Qualitative examples for other subsets are presented in the supplementary material.

\subsection{Train and Test Splits}

\subsubsection{SkyReg-Train.}
Our training split combines two cities from the Urban subset (Seattle and San Francisco) with all scenes from the Landmarks subset. Urban scenes span a wide range of buildings, streets, and landscapes across both cities. Data acquisition in this subset follows a relatively regular sampling pattern, pairing orthorectified satellite tiles with drone images captured at fixed altitudes and yaw angles. This split is designed to expose models to a large number of distinct environments and to encourage robustness to the substantial viewpoint and scale gap between orthographic satellite imagery and perspective drone views captured closer to the ground.

In contrast, the Landmarks subset contains fewer scenes but exhibits greater viewpoint diversity. For each landmark scene, we randomize both drone and satellite camera intrinsics and extrinsics, producing a broad distribution of camera configurations and relative poses. Together, the Urban and Landmarks subsets are complementary: the former provides breadth across many scenes under controlled sensor assumptions, while the latter provides challenging geometric variation, enabling models trained on SkyReg-Train to generalize across scene layouts, acquisition conditions, and camera models.

\vspace{-1em}

\subsubsection{SkyReg-Bench.}
SkyReg-Bench is created to evaluate geo-registration under strict geographic generalization, using images from entirely unseen locations and covering a broad range of environments, sensor models, and scene layouts. It combines a held-out city from the Urban subset (Chicago) with all scenes from the Suburban subset, yielding a challenging test set that differs substantially from the training distribution. The Urban subset is designed to evaluate robustness to new city-scale appearance statistics and scene semantics under the same orthorectified satellite--to--drone setting, while the Suburban subset evaluates generalization across less structured environments and more diverse acquisition conditions. Together, these components complement one another: the Urban split measures transfer across dense man-made scenes at scale, whereas the Suburban split emphasizes robustness to distribution shifts in viewpoint, layout, and scene type, providing a comprehensive benchmark for pixel-level cross-view geo-registration.

Table~\ref{tab:skyreg_stats} summarizes the dataset composition. We report the number of \emph{images} for both drone and satellite for each subset, how the \emph{depth} is derived, and the type of \emph{satellite projection}.

\begin{table}[t]
    \centering
    \caption{\textbf{SkyReg dataset statistics.} Number of drone queries and satellite references for each split and subset, along with the satellite projection type and the depth source used to derive dense geodetic supervision (LiDAR for Urban; SfM for Landmarks/Suburban).}
    \label{tab:skyreg_stats}
    \setlength{\tabcolsep}{4pt}
    \begin{tabular}{l l l l r r}
        \toprule
        \textbf{Split} & \textbf{Subset} & \textbf{Sat. Proj.} & \textbf{Depth} & \textbf{\#Drone} & \textbf{\#Sat.} \\
        \midrule
        \multirow{2}{*}{Train} 
            & Urban (SEA \& SF)& Orthographic & LiDAR & 27,836 & 152 \\
            & Landmarks & Perspective & SfM & 87,778 & 7,766 \\
        \midrule
        \multirow{2}{*}{Test} 
            & Urban (CHI) & Orthographic & LiDAR & 13,191 & 238 \\
            & Suburban & Perspective & SfM & 950 & 5 \\
        \bottomrule
    \end{tabular}
\end{table}

\subsection{Data Collection Details}

\subsubsection{Urban Scenes.}

We sample 390 orthorectified satellite tiles over Chicago, Seattle, and San Francisco. Reference images are taken from Bing Maps RGB tiles with side lengths of 380--750\,m. Query drone views are rendered in Google Earth with fixed intrinsics and extrinsics: cameras are placed 100\,m above ground, pitched at $45^\circ$, and assigned a random yaw. Viewpoints are selected by sampling GPS coordinates from VIGOR \cite{zhu2021vigor} and setting the camera pose so that the principal ray intersects the ground plane at the sampled location.

For depth supervision, we download geo-referenced LiDAR point clouds for each tile from USGS 3DEP \cite{Sugarbaker2014_3DEP_CIR1399}, expressed in the global ECEF frame. We form $DEM(\mathbf{x}_r)$ for each reference tile via interpolation from the LiDAR elevations. To obtain dense drone depth-maps, we first note that a world point $\tilde{X}=[X^\top,1]^\top$ projects to the query as $\tilde{\mathbf{x}}_q \sim K_q P^{w\to q}\tilde{X}$, where
\(
P^{w\to q}=[(R^{q\to w})^\top \;|\; -(R^{q\to w})^\top \mathbf{t}^{q\to w}]
\)
and $\tilde{\mathbf{x}}_q=[u,v,1]^\top$. Because direct projection yields sparse samples, we densify the LiDAR into a triangular mesh $\mathcal{M}$ and compute per-pixel intersections by ray tracing:
\begin{equation}
\label{eq:raytracing}
\mathbf{r}_q(s)=\mathbf{t}^{q\to w}+s\,R^{q\to w}K_q^{-1}\tilde{\mathbf{x}}_q,
\qquad
X_q(\mathbf{x}_q)=\mathrm{Intersect}\!\bigl(\mathbf{r}_q(\cdot),\mathcal{M}\bigr).
\end{equation}
The query depth map $D(\mathbf{x}_q)$ is then obtained from the intersection point (its depth along the camera ray).

\vspace{-1em}

\subsubsection{Landmark Scenes.}

We build the Landmarks subset from AerialMegaDepth scenes \cite{vuong2025aerialmegadepth}, which provide drone images with depth and camera parameters, and augment each scene with perspective satellite views rendered in Google Earth Studio. For each scene, we compute the convex hull of drone camera centers projected onto the ground plane, then sample 100 ground-plane anchors $\mathcal{A}$ uniformly inside the hull and up to 100 satellite camera centers $\mathcal{B}$ by sampling horizontal locations within the hull and altitudes up to 2500\,m. We pair points in $\mathcal{A}$ and $\mathcal{B}$, place a virtual satellite camera at each $\mathbf{b}\!\in\!\mathcal{B}$, and orient it toward its paired anchor in $\mathcal{A}$. We further randomize the field of view in {$[10^\circ, 45^\circ]$} to obtain diverse intrinsics and viewpoints.

Since these scenes span worldwide locations where LiDAR is often unavailable, we generate depth via a COLMAP SfM/MVS pipeline \cite{schoenberger2016sfm, schoenberger2016mvs}: SfM recovers metric-scale point clouds (with bundle adjustment), and MVS produces dense per-view metric depth maps. In total, Landmarks contains 7.7k satellite and {87k} drone perspective images with depth, 6-DoF poses, and intrinsics.

\subsubsection{Suburban Scenes.}

The Suburban scenes consist of {5} scenes from the WRIVA Public Dataset \cite{wrivapublic} with {950} drone images. This subset is dominated by non-urban environments and areas affected by natural disasters, yielding visual characteristics and scene structure that differ substantially from the training distribution. As a result, it serves as a challenging testbed for evaluating robustness and generalization to previously unseen settings. Both satellite and drone imagery in the Suburban setting use perspective projection and are collected similarly to the Landmarks subset. Specifically, we adopt the GPS spatial distribution provided by the WRIVA Public Dataset \cite{wrivapublic} to select acquisition locations, and then collect corresponding views from Google Earth with full camera parameter annotations. Finally, we run an SfM pipeline to reconstruct scene geometry and generate dense depth maps for the satellite and drone views.

\section{Benchmarking and Evaluation}

\subsection{Evaluation Protocol}

At test time, we feed the query and reference images into the SkyReg model and obtain the predicted scene geometry, including dense 3D point maps and the camera parameters required to define the cross-view warping function from Eq. \ref{eq:warp}. Using this function, we map each valid query pixel into the reference image plane and transfer the GPS coordinates to the query view by interpolating the geodetic map of the reference at the warped query pixel locations, yielding a dense set of predicted GPS coordinates for the query image.

We propose two complementary ways to quantify the model's performance. The first is a \emph{dense} metric that evaluates pixel-level geo-registration accuracy. For each query, after warping the query into the reference frame, we compare the predicted GPS coordinate of each valid query pixel against its ground-truth coordinate and find the \emph{geodetic error} (GE). Specifically, we compute the ground-truth latitude and longitude of each valid query pixel from the known depth maps and camera parameters. This yields, for each image, a set of pixels with predicted and ground-truth GPS coordinates. We then compute the per-pixel error using the Haversine distance between the predicted and ground-truth latitude--longitude pairs, from which we assign the median error as the geodetic error. Given two coordinates $(\phi_1,\lambda_1)$ and $(\phi_2,\lambda_2)$, the Haversine distance is
\begin{equation}
d_{\mathrm{hav}}
=
2R \arcsin\!\left(
\sqrt{
\sin^2\!\left(\frac{\phi_2-\phi_1}{2}\right)
+
\cos(\phi_1)\cos(\phi_2)\sin^2\!\left(\frac{\lambda_2-\lambda_1}{2}\right)
}
\right),
\end{equation}
where $R$ is the Earth's radius, and all angles are expressed in radians.

The second metric measures geo-localization performance. For each query, we compute a \emph{single} GPS coordinate to facilitate comparison with image-level cross-view geo-localization methods. Concretely, we compute an image-level GPS estimate by finding the median of the predicted pixel-wise coordinates over all valid pixels, and do the same for the ground-truth coordinates. We then report the geodetic distance between these two image-level estimates. This protocol mirrors retrieval baselines, where a query location is typically taken from a single point (often the center) of the retrieved reference. 


Finally, some baselines may fail catastrophically on a subset of samples and produce no valid registration. To measure robustness, we also report \emph{recall}---the fraction of query images that are successfully geo-registered under multiple distance thresholds. An image is considered correctly registered at threshold $\tau$ if its geodetic error is below $\tau$ m.

\subsection{Results and Discussion}

Tables \ref{tab:skyreg_georef} and \ref{tab:skyreg_georef_center} show that our method consistently outperforms retrieval-based and homography-based baselines by a large margin in both the pixel-wise and median-GPS mean geodetic error and recall. Relative to the strongest competing baseline (RoMa), which uses dense point correspondence, we reduce the pixel-wise mean geodetic error by 88.18 m and 90.76 m in the Urban and Suburban splits, respectively. On the median-GPS evaluation we observe a similar trend, reporting 68.30 m and 76.33 m lower mean geodetic error than RoMa. RoMa Dense, which uses the dense point correspondence from RoMa, has similar performance. Recall exhibits the same trend: SkyReg registers a substantially larger fraction of images across all distance thresholds, indicating robust behavior rather than gains concentrated on a small subset of easy examples.

The improvements are most pronounced on the Urban subset, which is particularly challenging due to tall structures and strong parallax. In these scenes, planar homography assumptions often break down and can lead to catastrophic mis-registrations, while our 3D-aware formulation remains stable. Retrieval methods perform worst overall, as they inherit the discretization and viewpoint mismatch of the gallery; notably, even strong matchers such as SP+SG can underperform simpler cross-view retrieval baselines (e.g., University-1652) in the Urban setting. RoMa is generally the overall best-performing non-3D baseline, but it still falls well short of SkyReg, underscoring the benefit of explicitly modeling scene geometry for dense cross-view geo-registration.

\begin{table}[t]
\caption{Pixel-wise georegistration performance on the Urban and Suburban benchmarks. We report the mean geodetic error (GE) in meters, and the recall of images registered correctly under $\tau$ as a percentage.}
\label{tab:skyreg_georef}
\resizebox{\linewidth}{!}{
\begin{tabular}{l|ccccc|ccccc}
\hline \hline
\multicolumn{1}{c|}{\multirow{3}{*}{\textbf{Method}}} & \multicolumn{5}{c|}{\textbf{Urban}}                                                                             & \multicolumn{5}{c}{\textbf{Suburban}}                                                                           \\ \cline{2-11} 
\multicolumn{1}{c|}{}                                 & \multicolumn{1}{c|}{\multirow{2}{*}{\textbf{GE} $\downarrow$}} & \multicolumn{4}{c|}{\textbf{Recall@$\tau$ (m)} $\uparrow$}                         & \multicolumn{1}{c|}{\multirow{2}{*}{\textbf{GE} $\downarrow$}} & \multicolumn{4}{c}{\textbf{Recall@$\tau$ (m)} $\uparrow$}                          \\
\multicolumn{1}{c|}{}                                 & \multicolumn{1}{c|}{}                              & \textbf{20}    & \textbf{30}    & \textbf{40}    & \textbf{50}    & \multicolumn{1}{c|}{}                              & \textbf{20}    & \textbf{30}    & \textbf{40}    & \textbf{50}    \\ \hline
SP+SG \cite{detone2018superpoint, sarlin2020superglue}                                                & \multicolumn{1}{c|}{209.40}                        & 1.40           & 1.94            & 2.34           & 2.82          & \multicolumn{1}{c|}{223.69}                        & 36.42          & 37.15          & 37.47          & 37.47          \\
RoMa \cite{edstedt2024roma}                                                 & \multicolumn{1}{c|}{133.06}                        & 25.69          & 27.18          & 28.35          & 29.33          & \multicolumn{1}{c|}{112.10}                         & 60.00          & 61.26          & 62.25          & 63.15          \\ 
RoMa Dense \cite{edstedt2024roma}                                                 & \multicolumn{1}{c|}{131.45}                        & 17.62          & 20.08          & 22.43          & 24.72          & \multicolumn{1}{c|}{111.25}                         & 51.89          & 52.52          & 53.57         & 54.42 \\

\hline
\textbf{Ours}                                         & \multicolumn{1}{c|}{\textbf{44.88}}                & \textbf{78.78} & \textbf{80.03} & \textbf{80.53} & \textbf{80.94} & \multicolumn{1}{c|}{\textbf{21.34}}                & \textbf{74.94} & \textbf{93.57} & \textbf{97.78} & \textbf{98.52} \\ \hline \hline
\end{tabular}
}
\end{table}

\begin{table}[t]

\caption{Median-GPS geo-localization performance on Urban and Suburban benchmarks. We report the mean geodetic error (GE) in meters, and the recall of images geolocated correctly under $\tau$ as a percentage.}
\label{tab:skyreg_georef_center}
\resizebox{\linewidth}{!}{
\begin{tabular}{l|ccccc|ccccc}
\hline \hline
\multicolumn{1}{c|}{\multirow{3}{*}{\textbf{Method}}} & \multicolumn{5}{c|}{\textbf{Urban}}                                                                             & \multicolumn{5}{c}{\textbf{Suburban}}                                                                           \\ \cline{2-11} 
\multicolumn{1}{c|}{}                                 & \multicolumn{1}{c|}{\multirow{2}{*}{\textbf{GE} $\downarrow$}} & \multicolumn{4}{c|}{\textbf{Recall@$\tau$ (m)} $\uparrow$}                         & \multicolumn{1}{c|}{\multirow{2}{*}{\textbf{GE} $\downarrow$}} & \multicolumn{4}{c}{\textbf{Recall@$\tau$ (m)} $\uparrow$}                          \\
\multicolumn{1}{c|}{}                                 & \multicolumn{1}{c|}{}                              & \textbf{20}    & \textbf{30}    & \textbf{40}    & \textbf{50}    & \multicolumn{1}{c|}{}                              & \textbf{20}    & \textbf{30}    & \textbf{40}    & \textbf{50}    \\ \hline
TransGeo \cite{zhu2022transgeo}                                             & \multicolumn{1}{c|}{160.98}                        & 1.50            & 3.47           & 5.98           & 9.06           & \multicolumn{1}{c|}{272.53}                        & 0.31           & 1.47           & 1.99           & 2.94           \\
University-1652 \cite{zheng2020university}                                      & \multicolumn{1}{c|}{121.17}                        & 4.25           & 9.06           & 15.01          & 21.48          & \multicolumn{1}{c|}{209.60}                         & 1.05           & 2.42           & 3.36           & 4.84           \\
EarthMatch \cite{berton2024earthmatch}                                           & \multicolumn{1}{c|}{163.60}                        & 1.80           & 3.83           & 6.57           & 9.13           & \multicolumn{1}{c|}{302.34}                              &     0.42           &      0.74          &      1.26          &   1.79             \\ \hline
SP+SG \cite{detone2018superpoint, sarlin2020superglue}                                                & \multicolumn{1}{c|}{134.09}                        & 1.97           & 3.60            & 5.76           & 8.07          & \multicolumn{1}{c|}{122.47}                        & 32.41          & 35.37          & 38.33          & 40.76          \\
RoMa \cite{edstedt2024roma}                                                 & \multicolumn{1}{c|}{104.70}                        & 21.75          & 26.42          & 29.34          & 31.77          & \multicolumn{1}{c|}{97.78}                         & 48.05          & 53.62         & 57.29          & 59.28         \\ 

RoMa Dense \cite{edstedt2024roma}                                                 & \multicolumn{1}{c|}{98.38}                        & 18.66          & 24.58          & 29.41          & 34.52          & \multicolumn{1}{c|}{86.91}                         & 40.29         & 45.54        & 48.68        & 52.25          \\ \hline
\textbf{Ours}                                         & \multicolumn{1}{c|}{\textbf{36.40}}                & \textbf{66.72} & \textbf{78.16} & \textbf{80.47} & \textbf{81.57} & \multicolumn{1}{c|}{\textbf{21.45}}                & \textbf{75.32} & \textbf{88.55} & \textbf{93.17} & \textbf{95.27} \\ \hline \hline
\end{tabular}
}
\end{table}

\subsection{Qualitative Results}
\begin{figure}[t]
    \centering
    \includegraphics[width=0.95\linewidth]{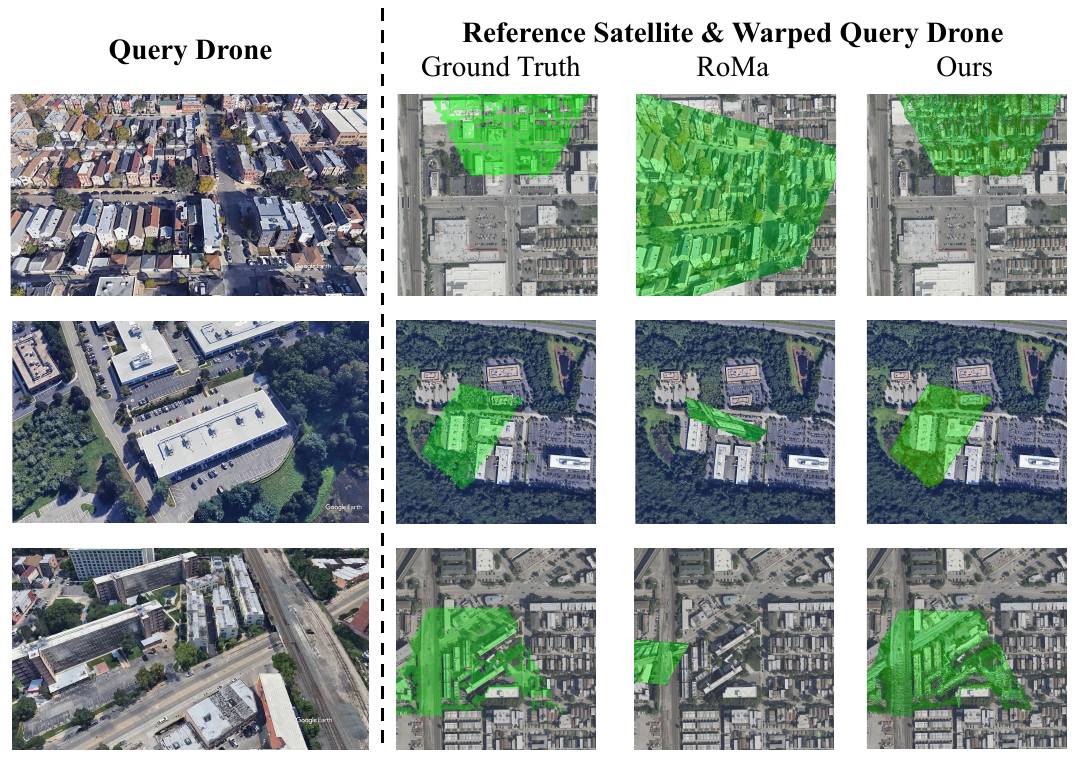}
    \caption{\textbf{Qualitative geo-registration on SkyReg-Bench.} For each example, we warp the query drone image into the geodetically accurate satellite reference frame and visualize the overlay in green. From left to right: (1) query drone image, (2) ground-truth warp obtained using the ground-truth depth map and camera parameters, (3) RoMa result (homography-based), and (4) SkyReg (ours). RoMa often fails under large cross-view viewpoint/scale changes; even when successful, a single planar homography cannot model parallax, leading to stretching or shearing of elevated structures. In contrast, SkyReg’s 3D-aware warping more closely matches the ground-truth alignment and naturally respects occlusions, producing sharper, geometrically consistent overlays.}
    \label{fig:pose_results}
\end{figure}

To provide additional intuition, Figure~\ref{fig:pose_results} shows three examples in which we warp the query drone image into the reference frame and overlay it (green) on the georeferenced satellite image (RGB). We compare against RoMa. In this cross-view setting, RoMa often fails to estimate a plausible homography, resulting in large geodetic errors and visibly misaligned overlays. Even when RoMa produces a reasonable registration, the planar warp can introduce noticeable distortions---most prominently stretching or shearing of elevated structures---because a single homography cannot explain parallax induced by height variations. In contrast, our 3D-aware model yields accurate alignments under extreme scale changes, large viewpoint differences, and partial occlusions.

\subsection{Training Ablation}

Here we perform an additional ablation by varying the training data used to
train the SkyReg model. Specifically, we evaluate three variants:

\begin{itemize}
\item \textbf{Landmark Subset} – SkyReg trained on Landmarks subset and 3D datasets.
\item \textbf{Urban Subset} – SkyReg trained on Urban subset and 3D datasets.
\item \textbf{Landmark+Urban} – SkyReg trained on combined Urban, Landmark and 3D datasets as described in Section~3.4.
\end{itemize}

All models are evaluated on the SkyReg-Bench benchmark using the same
evaluation protocol for median-GPS geo-localization described in Section~5.1.
\begin{table}[t]
\caption{Median-GPS geo-localization performance on Urban and Suburban benchmarks. We report the mean geodetic error (GE) in meters, and the recall of images geolocated correctly under $\tau$ as a percentage.}
\label{tab:skyreg_georef_center2}
\resizebox{\linewidth}{!}{
\begin{tabular}{l|ccccc|ccccc}
\hline \hline
\multicolumn{1}{c|}{\multirow{3}{*}{\textbf{Method}}} & \multicolumn{5}{c|}{\textbf{Urban}}                                                                             & \multicolumn{5}{c}{\textbf{Suburban}}                                                                           \\ \cline{2-11} 
\multicolumn{1}{c|}{}                                 & \multicolumn{1}{c|}{\multirow{2}{*}{\textbf{GE} $\downarrow$}} & \multicolumn{4}{c|}{\textbf{Recall@$\tau$ (m)} $\uparrow$}                         & \multicolumn{1}{c|}{\multirow{2}{*}{\textbf{GE} $\downarrow$}} & \multicolumn{4}{c}{\textbf{Recall@$\tau$ (m)} $\uparrow$}                          \\
\multicolumn{1}{c|}{}                                 & \multicolumn{1}{c|}{}                              & \textbf{20}    & \textbf{30}    & \textbf{40}    & \textbf{50}    & \multicolumn{1}{c|}{}                              & \textbf{20}    & \textbf{30}    & \textbf{40}    & \textbf{50}    \\ \hline


Landmark Subset                                                 & \multicolumn{1}{c|}{48.70}                        & 59.89           & 69.83            & 72.03           & 73.18          & \multicolumn{1}{c|}{27.23}                        & 73.66          & 85.20          & 89.93          & 91.40          \\
Urban Subset                                                & \multicolumn{1}{c|}{50.38}                        & 58.47          & 66.19          & 68.26         & 69.37          & \multicolumn{1}{c|}{24.79}                         & 72.30          & 86.46          & 91.92          & 93.07          \\ \hline
\textbf{Landmark+Urban}                                         & \multicolumn{1}{c|}{\textbf{36.40}}                & \textbf{66.72} & \textbf{78.16} & \textbf{80.47} & \textbf{81.57} & \multicolumn{1}{c|}{\textbf{21.45}}                & \textbf{75.32} & \textbf{88.55} & \textbf{93.17} & \textbf{95.27} \\ \hline \hline
\end{tabular}
}
\end{table}
These results (Table \ref{tab:skyreg_georef_center2}) highlight the importance of diverse training data for robust
cross-view geo-registration. Training on a single subset reduces
generalization performance, while combining Urban and Landmark scenes
provides the best overall accuracy.
\section{Conclusion}


We present \emph{SkyReg}, a geometry-aware approach to pixel-level drone--satellite geo-registration that leverages feed-forward 3D reconstruction to estimate the cross-view warping between a perspective drone query and a geodetically accurate satellite reference, enabling dense pixel-wise GPS assignment without relying on 2D point matches and remaining robust to occlusions and large viewpoint changes. To support training and rigorous evaluation, we introduced SkyReg-Train and SkyReg-Bench, providing dense per-pixel latitude–longitude supervision with depth and camera parameters derived from LiDAR and SfM, and testing generalization on unseen Urban and Suburban scenes. Across both dense and image-level protocols, SkyReg consistently outperforms strong retrieval and homography baselines, substantially improving geodetic error and recall (e.g., large gains over RoMa on unseen Urban/Suburban splits). These findings establish SkyReg as a reliable baseline for structure-level geodetic alignment and provide a foundation for future research on dense cross-view geo-registration.

\section*{Acknowledgments}
Supported by Intelligence Advanced Research Projects Activity (IARPA) via Department of Interior/Interior Business
Center (DOI/IBC) contract number 140D0423C0074. The
U.S. Government is authorized to reproduce and distribute
reprints for Governmental purposes notwithstanding any
copyright annotation thereon. Disclaimer: The views
and conclusions contained herein are those of the authors
and should not be interpreted as necessarily representing
the official policies or endorsements, either expressed or
implied, of IARPA, DOI/IBC, or the U.S. Government.

%
%
\bibliographystyle{splncs04}
\bibliography{main}

\clearpage
\appendix


\title{Pixel-wise Geo-registration of Drone Images (Supplementary Material)} 


\author{Qingyang Liu\textsuperscript{*} \quad David G. Shatwell\textsuperscript{*} \quad Parth Parag Kulkarni \quad Mubarak Shah}

\begingroup
\renewcommand{\thefootnote}{\fnsymbol{footnote}}
\footnotetext[1]{These authors contributed equally to this work.}
\endgroup

\authorrunning{Q.~Liu et al.}

\institute{
Institute of Artificial Intelligence, University of Central Florida\\
}

\maketitle
\vspace{0.5cm}

In this supplementary material, we provide additional details 
on our method and dataset. 
We organize our supplementary material into the following sections: 
\vspace{0.2in}
\begin{enumerate}[A.]
\setlength\itemsep{1.5em}
    \item Additional Data Gathering Details
    \vspace{0.15in}
    \begin{enumerate}[label=A.\arabic*.]
\setlength\itemsep{1.5em}
        \item Urban Scenes
        \item Landmarks and Suburban Scenes
    \end{enumerate}
    \item Additional Dataset Information
    \vspace{0.15in}
    \begin{enumerate}[label=B.\arabic*.] \setlength\itemsep{1.5em}
        \item SkyReg Urban
        \item SkyReg Landmark
        \item SkyReg Suburban
    \end{enumerate}
    \item Additional Qualitative Results
\end{enumerate}

\section{Additional Data Gathering Details}
\subsection{Urban Scenes}

\subsubsection{Ground Truth Pixel-wise GPS}
For the Urban subset of SkyReg, the ground-truth geodetic data for each reference tile is obtained from USGS 3DEP dataset with the official TNM Download V2 tool in the form of LiDAR files, each containing approximately 18 million 3D points in each tile’s native projected coordinate system. The planar coordinates $(x, y)$ are transformed into GPS coordinates using a coordinate reference system conversion to WGS--84, yielding latitude and longitude values. The elevation coordinate $z$, after unit conversion when necessary, is used as the altitude. This process produces a georeferenced point map in which each location is represented by its corresponding $(\text{lat}, \text{lon}, \text{alt})$ values, which serve as the ground truth for the satellite reference tiles.

\subsubsection{Reference Satellite Images}
To obtain the corresponding RGB reference satellite images, we use the GPS coordinates of the four corners of each LiDAR tile to retrieve satellite imagery from Bing Maps. We choose Bing Maps because its satellite images are orthorectified, which allows us to use the LiDAR point clouds to project and estimate depth at each pixel without requiring the ground-truth satellite pose. The images are collected at zoom level 17 for Chicago and 16 for Seattle and San Francisco since the latter two encompass a significantly larger area, providing high spatial resolution. Since the Bing Maps imagery is orthorectified and the GPS coordinates of the four tile corners are known, we can interpolate the geographic coordinates of all remaining pixels, thereby obtaining a complete georeferencing of the reference satellite images.

\subsubsection{Query Drone Images}
For the queries, we define synthetic RGB images along with camera intrinsics and poses expressed in the ECEF coordinate system. To compute ground-truth geodetic coordinates for each valid pixel, we use LiDAR points from the corresponding reference tile and its neighboring tiles, transform them to ECEF coordinates, and project them into the query camera frame. The camera intrinsics are then used to determine the correspondence between projected LiDAR points and query pixels, so that each matched pixel can be assigned an ECEF coordinate and subsequently converted to geodetic coordinates.

Since the query images are captured much closer to the ground, they exhibit higher spatial resolution than the reference data. Consequently, this direct projection process yields only a sparse ground-truth point map: large portions of the image remain unlabeled, and some pixel assignments may be incorrect due to occlusions or projections onto geometry behind foreground structures. To overcome this limitation, we construct a surface mesh using 2D Delaunay triangulation and then perform ray tracing over the mesh to obtain a dense depth map in metric scale (Eq. \ref{eq:raytracing} in main paper). Using the known camera pose and intrinsics, we convert this depth map into a dense point map in ECEF coordinates, which is then transformed into geodetic coordinates. This procedure significantly densifies the ground-truth map and allows nearly every pixel in the query image to be assigned an accurate geodetic coordinate. To keep mesh triangulation computationally tractable, we sample 300k points for mesh construction and enforce a maximum mesh edge length of 20 meters to prevent the mesh from spanning large empty regions.

\subsection{Landmarks and Suburban Scenes}

For landmarks and suburban scenes, we first define a set of camera poses and intrinsics over a predefined area, following the procedure described in Section 4.2 of the main paper. Once the images, camera poses, and intrinsics are available, we keep the camera parameters fixed and run triangulation to recover the 3D scene structure. In this process, pairs of spatially nearby images are selected, local features are extracted and matched across those pairs, and the resulting correspondences are used together with the known camera geometry to estimate 3D points in the scene. In other words, triangulation computes the 3D location of scene points by intersecting the viewing rays associated with matching image observations from multiple cameras. Since the camera poses and intrinsics are fixed, this step focuses on reconstructing the scene geometry rather than estimating the camera parameters themselves. The reconstructed 3D points can then be projected back into each image to obtain per-image scene geometry, and the final output is a set of metric depth maps for each image in the scene.

\begin{figure}[!ht]
    \centering
    \includegraphics[width=0.95\linewidth]{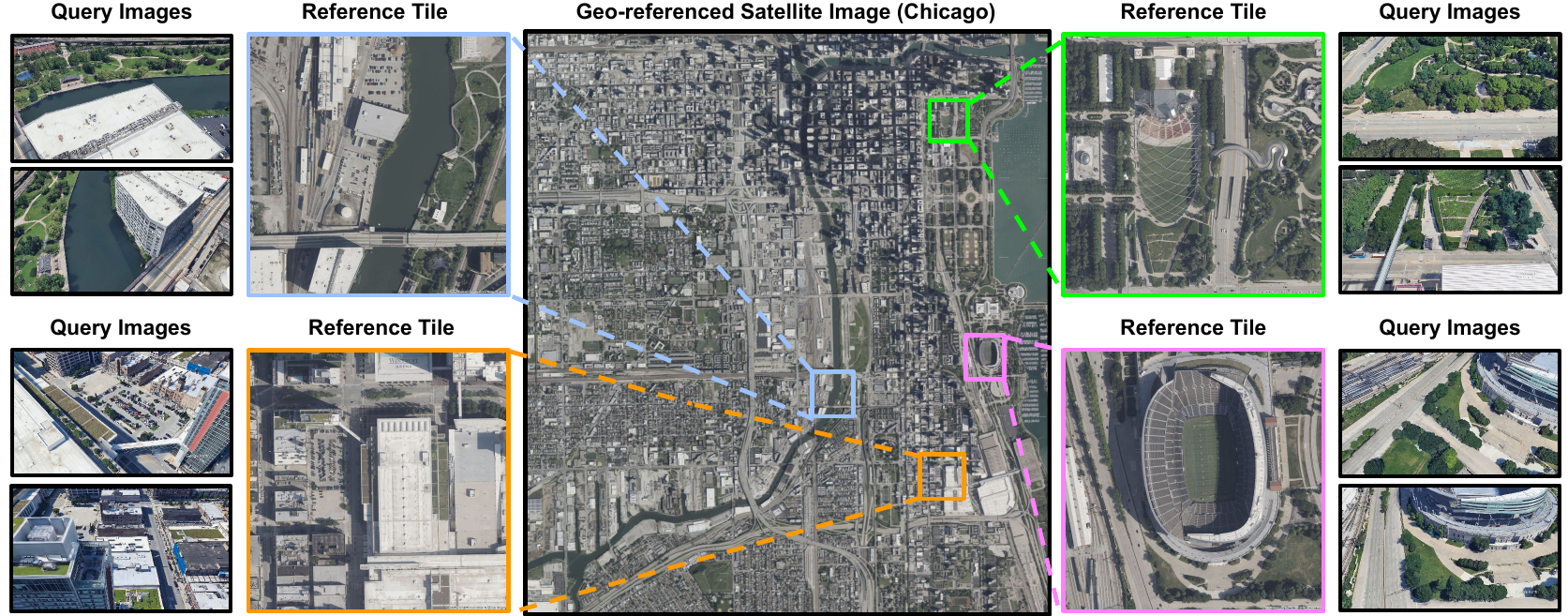}
    \caption{\textbf{Representative samples from SkyReg Urban Chicago.} The center panel shows a large orthorectified satellite image of Chicago. We create reference satellite images from local reference tiles. The left/right panels show example drone queries associated with each reference tile.}
    \label{fig:dataset_examples_chicago}
\end{figure}

\vspace{-0.4in}

\begin{figure}[!h]
    \centering
    \includegraphics[width=0.95\linewidth]{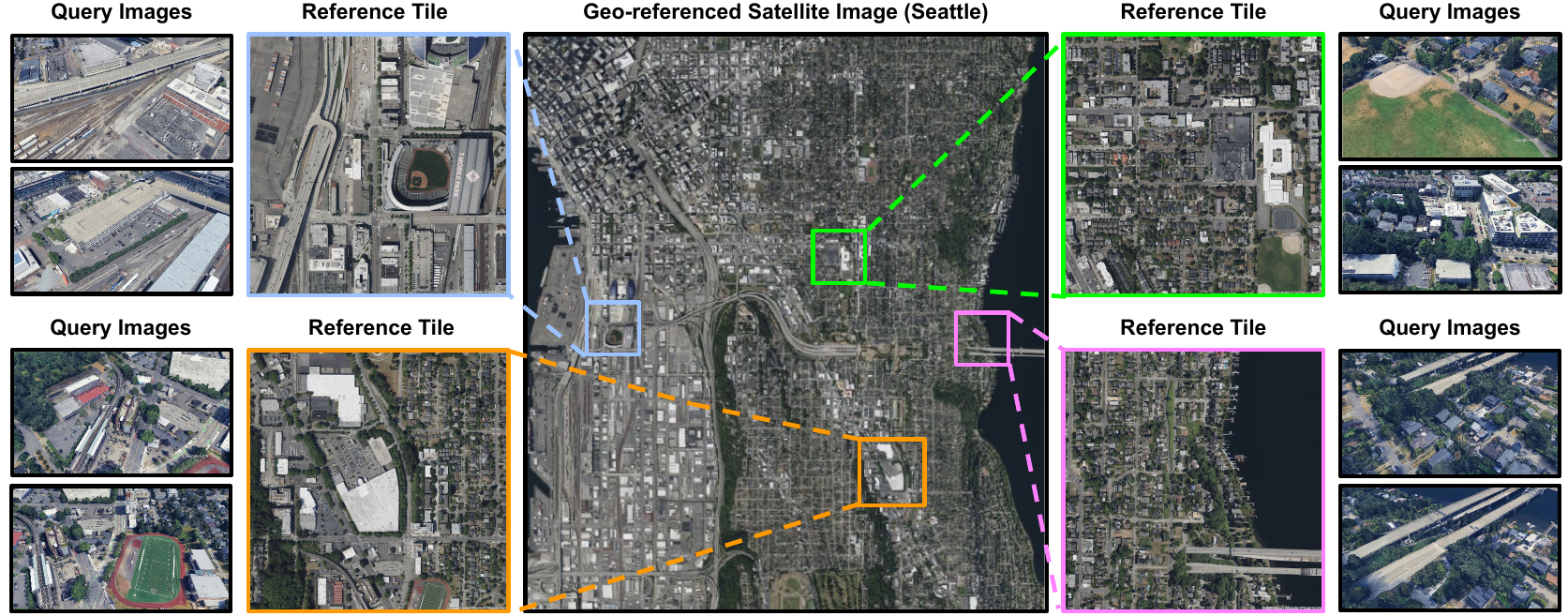}
    \caption{\textbf{Representative samples from SkyReg Urban Seattle.} The center panel shows a large orthorectified satellite image of Seattle. We create reference satellite images from local reference tiles. The left/right panels show example drone queries associated with each reference tile.}
    \label{fig:dataset_examples_seattle}
\end{figure}

\section{Additional Dataset Information} 
\label{sec:supp_distribution} 

We provide additional statistics describing the distribution of the dataset introduced in Section~4 of the main paper.

\subsection{SkyReg Urban}
SkyReg Urban consists of three cities-- Seattle, San Francisco, and Chicago (distribution shown in Table \ref{tab:city_distribution}).  Combined, the three cities consist of 41,027 drone images and 390 satellite images. Figures~\ref{fig:dataset_examples_chicago} and \ref{fig:dataset_examples_seattle} show sample reference and drone images for the cities of Chicago and Seattle respectively.

\vspace{-0.25in}
\begin{table}[h]
\centering
\caption{Number of drone and satellite images for each city in SkyReg Urban subset.}
\label{tab:city_distribution}
\vspace{-0.15in}
\begin{tabular}{lcc}
\toprule
\textbf{City} & \textbf{\# Drone} & \textbf{\# Sat.} \\
\midrule
Chicago & 13,191 & 238 \\
San Francisco & 14,346 & 71 \\
Seattle & 13,490 & 81 \\
\bottomrule
\end{tabular}
\end{table}

\vspace{-0.4in}

\subsection {SkyReg Landmark}
\begin{figure}[!h]
    \centering
    \includegraphics[width=\linewidth]{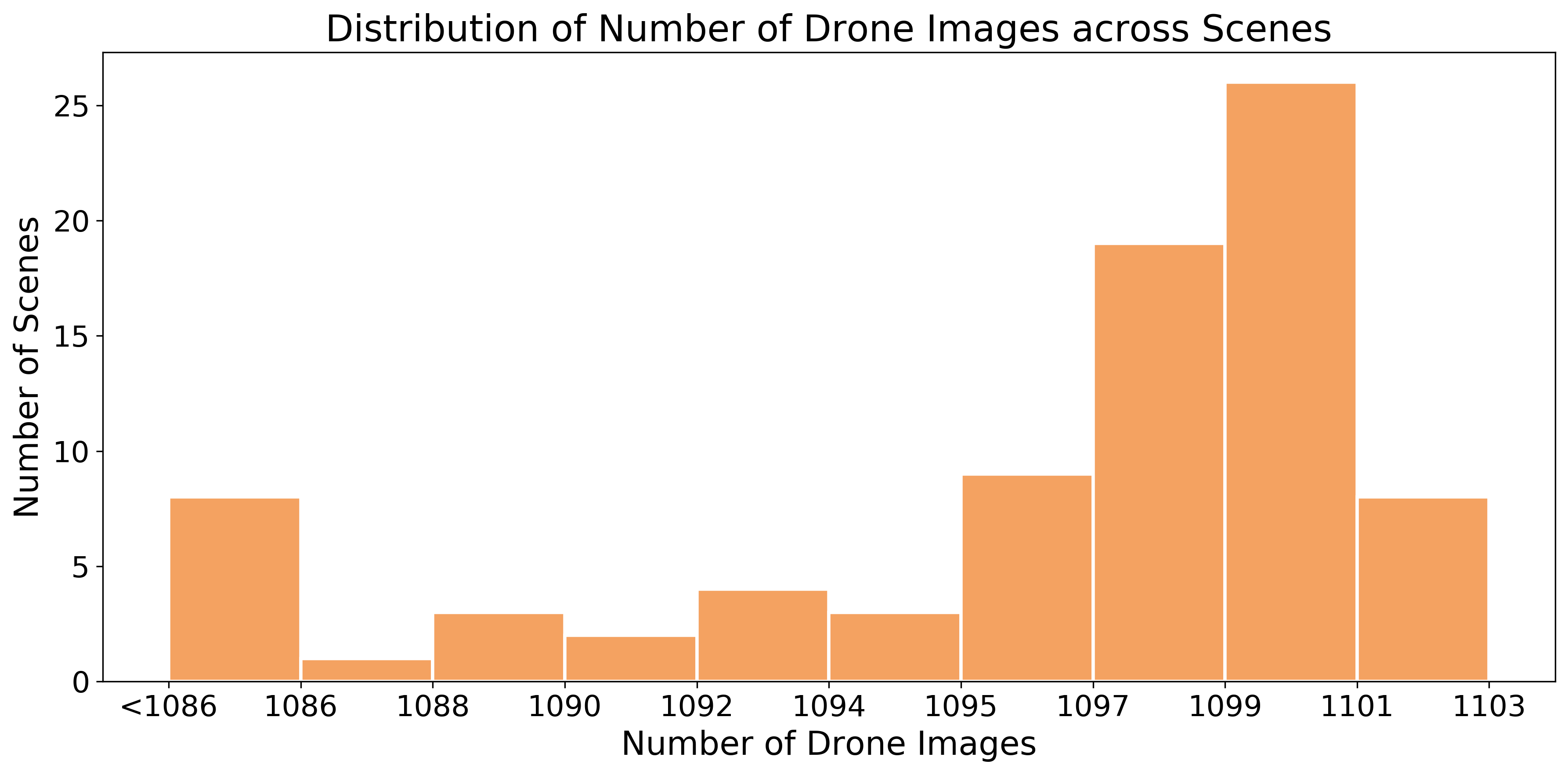}
    \caption{\textbf{Histogram showing the distribution of number of drone images across scenes for SkyReg Landmark}  Most scenes contain between approximately 1.0k and 1.1k drone images. This indicates that the dataset provides dense image coverage for majority of scenes. Freedman–Diaconis rule was used to compute the number of bins.}
    \label{fig:histogram}
\end{figure}

\begin{figure}[!ht]
    \centering
    \includegraphics[width=\linewidth]{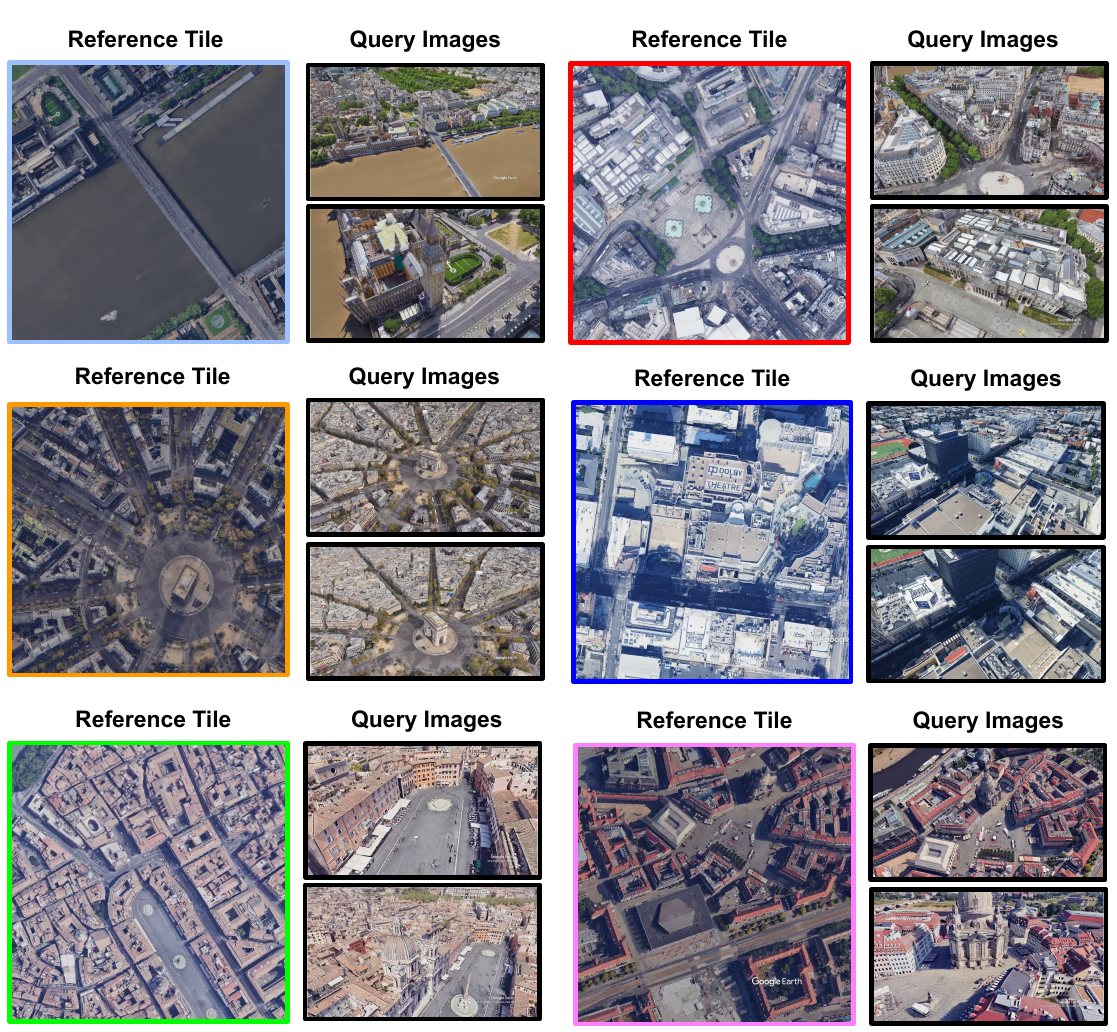}
    \caption{\textbf{Representative samples from SkyReg Landmark} Each pair consists of a drone query image and the reference tile consists of a satellite image centered around a significant landmark.}
    \label{fig:dataset_examples_landmarks}
\end{figure}

SkyReg Landmark consists of 83 scenes, each centered around landmarks located around the world. We show a histogram of the number of drone images per scene in Figure \ref{fig:histogram}. Figure~\ref{fig:dataset_examples_landmarks} shows representative samples from SkyReg Landmark. 

\vspace{-0.5in}
\subsection {SkyReg Suburban}

\begin{figure}[!ht]
    \centering
    \includegraphics[width=0.9\linewidth]{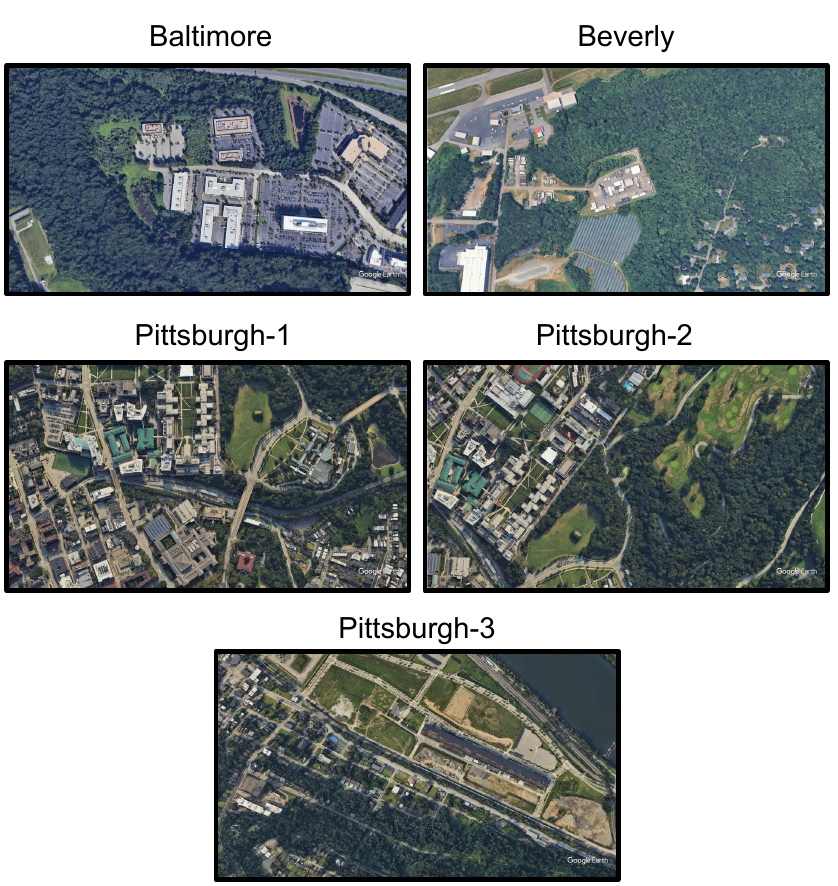}
    \caption{\textbf{Satellite Reference Tile from SkyReg Suburban} SkyReg Suburban consists of 5 scenes, each one consisting of one satellite image encompassing the entire scene.}
    \label{fig:dataset_examples_suburban}
\end{figure}

The Suburban split of SkyReg-Bench comprises five scenes across three U.S. states: one scene from Baltimore, MD; one from Beverly, MA; and three from Pittsburgh, PA. Each scene contains between 100 and 250 drone query images, each associated with a georeferenced satellite image and corresponding ground-truth GPS coordinates. The exact data distribution is shown in Table \ref{tab:suburban-dist}. Figure~\ref{fig:dataset_examples_suburban} shows sample reference images for each of the suburban scenes.

\begin{table}[h]
\centering
\caption{SkyReg Suburban number of drone images for each scene.}
\label{tab:suburban-dist}

\begin{tabular}{lc}
\toprule
\textbf{Scene} & \textbf{\# Images} \\
\midrule
Baltimore & 100 \\
Beverly & 100 \\
Pittsburgh-1 & 250 \\
Pittsburgh-2 & 250 \\
Pittsburgh-3 & 250 \\
\bottomrule
\end{tabular}

\end{table}

\newpage
\section{Additional Qualitative Results}
\label{sec:supp_qual}

\begin{figure*}[!ht]
\centering

\includegraphics[width=\textwidth]{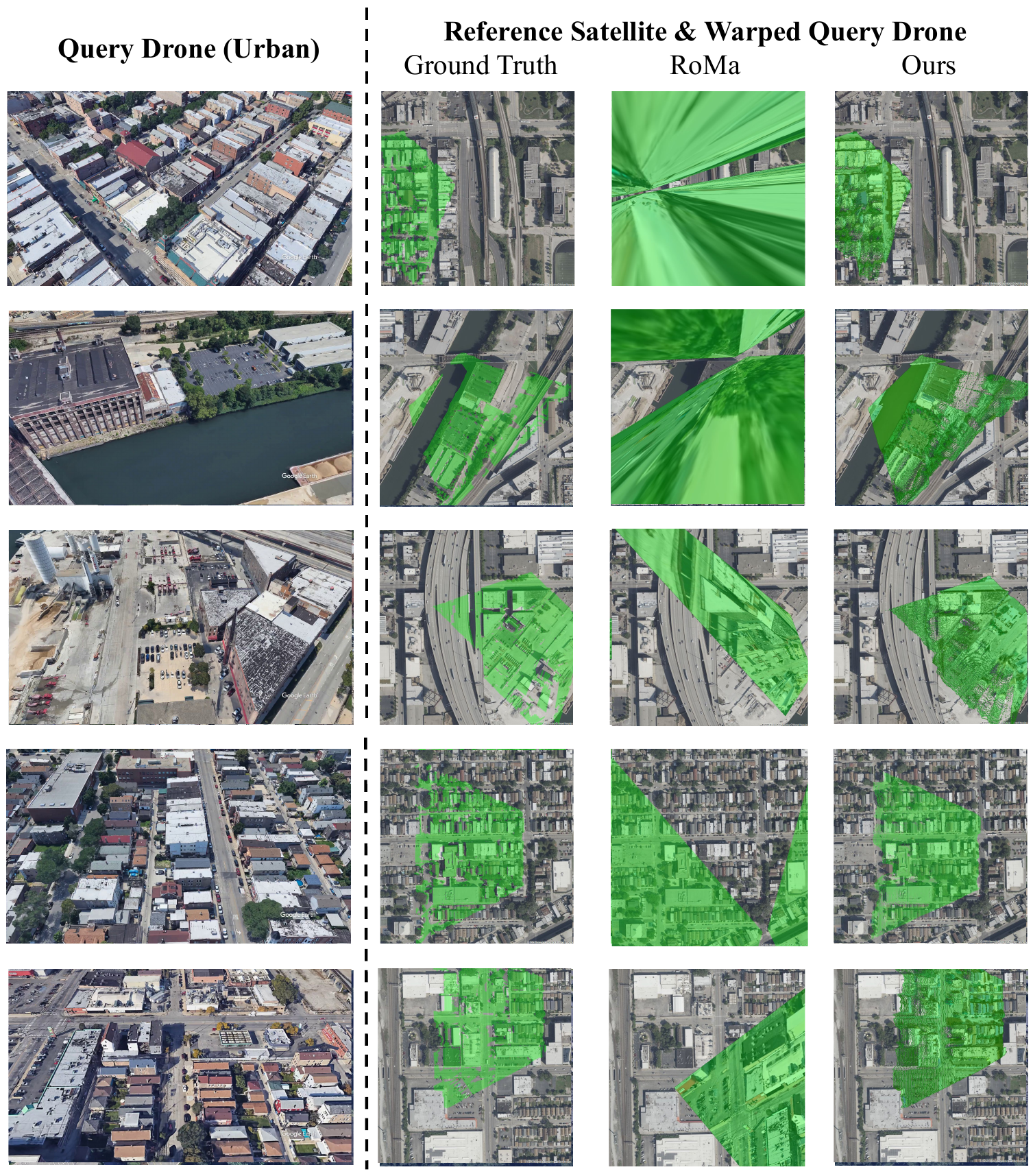}

\caption{
Additional qualitative examples of geo-registration on the Urban subset of SkyReg-Bench.
From left to right: (1) query drone image, (2) ground-truth warp,
(3) RoMa result, and (4) SkyReg (ours). The warped query image is
visualized in green on top of the georeferenced satellite reference.
}
\label{fig:supp_qualitative_1}

\end{figure*}

\begin{figure*}[!ht]
\centering

\includegraphics[width=\textwidth]{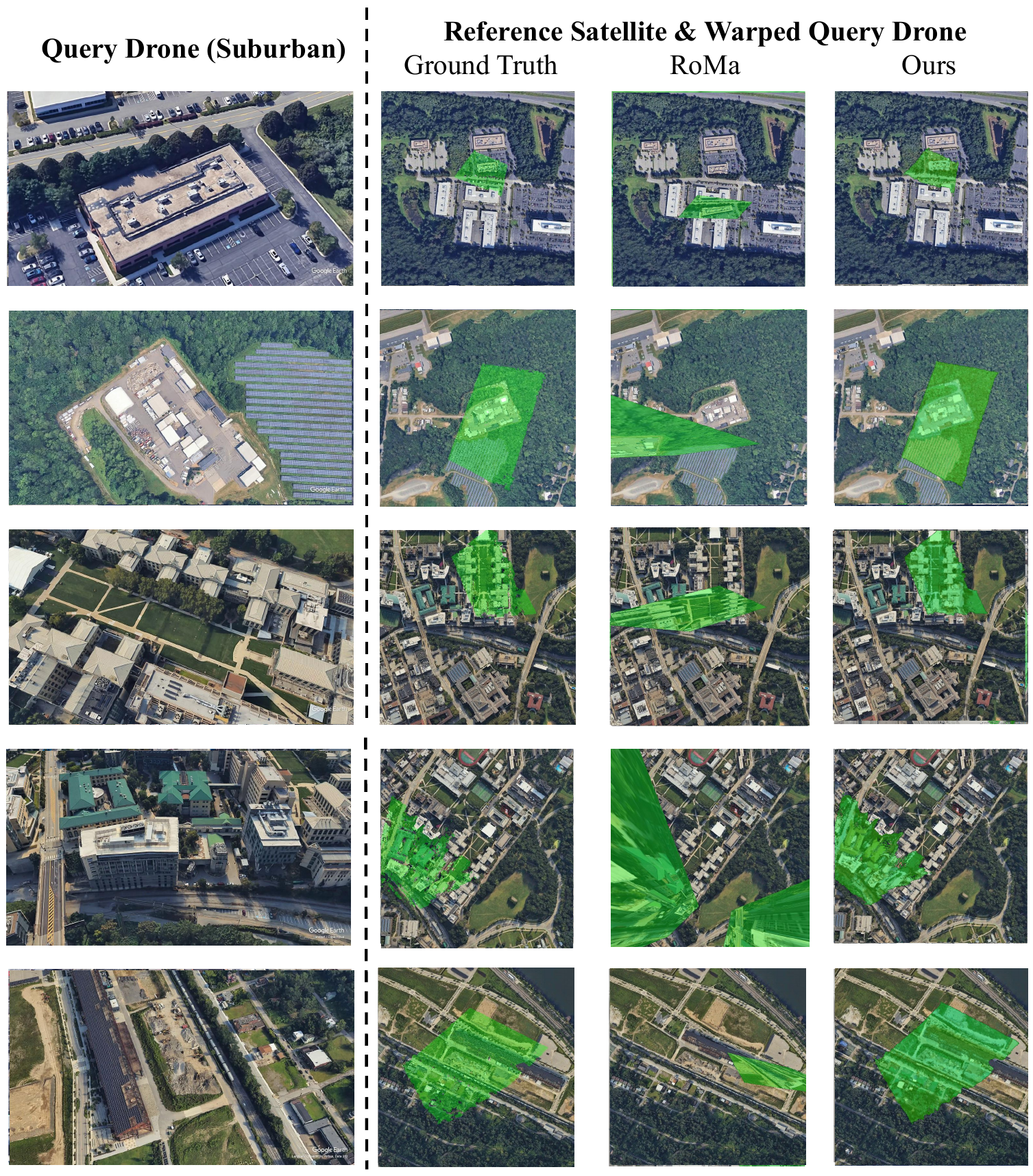}

\caption{
Additional qualitative examples of geo-registration on the Suburban subset of SkyReg-Bench.
From left to right: (1) query drone image, (2) ground-truth warp,
(3) RoMa result, and (4) SkyReg (ours). The warped query image is
visualized in green on top of the georeferenced satellite reference.
}
\label{fig:supp_qualitative_2}

\end{figure*}

Figures~\ref{fig:supp_qualitative_1} and \ref{fig:supp_qualitative_2} provide additional qualitative results
on the SkyReg-Bench dataset. Similar to Figure~3 in the main paper, we
visualize the warped query image overlaid on the georeferenced satellite
reference image. The overlays are shown in green to highlight the alignment.

These examples further demonstrate that the proposed geometry-aware
registration produces accurate alignments even under large viewpoint
differences and significant scene depth variation.

\end{document}